\documentclass[11pt,a4paper]{article}
\usepackage[hyperref]{paper}
\usepackage{times}
\usepackage{latexsym}
\usepackage{graphicx}
\usepackage{amsfonts}
\usepackage{multirow}
\usepackage{amsmath}
\usepackage{mathtools}

\usepackage{url}

\aclfinalcopy 

\title{Federated Learning Of Out-Of-Vocabulary Words}

\author{Mingqing Chen\qquad Rajiv Mathews\qquad Tom Ouyang\qquad Fran{\c{c}}oise Beaufays\\
Google LLC \\
  Mountain View, CA, U.S.A. \\
  {\tt {mingqing, mathews, ouyang, fsb}@google.com}}

\date{}

\begin{document}
\maketitle
\begin{abstract}
We demonstrate that a character-level recurrent neural network is able to learn
out-of-vocabulary (OOV) words under federated learning settings, for the purpose of expanding the vocabulary of
a virtual keyboard for smartphones without exporting sensitive text to servers.
High-frequency words can be sampled from the trained generative model
by drawing from the joint posterior directly.
We study the feasibility of the approach in two settings:
(1) using simulated federated learning on a publicly available non-IID per-user dataset from a popular social networking website,
(2) using federated learning on data hosted on user mobile devices.
The model achieves good recall and precision compared to ground-truth OOV words in setting (1).
With (2) we demonstrate the practicality of this approach by showing that
we can learn meaningful OOV words with good character-level prediction accuracy and cross entropy loss.

\end{abstract}

\section{Introduction}
Gboard --- the Google keyboard --- is a virtual keyboard for touch screen mobile
devices with support for more than 600 language varieties and over 1 billion installs as of 2018.
Gboard provides a variety of input features, including
tap and word-gesture typing, auto-correction, word completion and next word prediction.

Learning frequently typed words from user-generated data is an important component to the development of a mobile keyboard.
Example usage includes incorporating new trending words (celebrity names, pop culture words, etc.)
as they arise, or simply compensating for omissions in the initial keyboard implementation,
especially for low-resource languages.

The list of words is often referred to as the ``vocabulary'', and may or may not be hand-curated.
Words missing from the vocabulary cannot be predicted on the keyboard suggestion strip,
cannot be gesture typed, and more annoyingly, may be autocorrected even when typed correctly~\cite{ouyang2017mobile}.
Moreover, for latency and reliability reasons,
mobile keyboards models run on-device.
This means that the vocabulary supporting the models are intrinsically limited in size,
e.g. to a couple hundred thousand words per language.
It is therefore crucial to discover and include the most useful words in this rather short vocabulary list.
Words not in the vocabulary are often called ``out-of-vocabulary" (OOV) words.
Note that the concept of vocabulary is not limited to mobile keyboards.
Other natural language applications, such as for example neural machine translation (NMT),
rely on a vocabulary to encode words during end-to-end training.
Learning OOV words and their rankings is thus a fairly generic technology need.

The focus of our work is learning OOV words in a environment without transmitting and storing sensitive user content on centralized servers.
Privacy is easier to ensure when words are learned on device.
Our work builds upon recent advances in Federated Learning (FL)~\cite{konevcny2016federated, tim2018, hard2018federated},
a decentralized learning technique to train models such as neural networks on users' devices,
uploading only ephemeral model updates to the server for aggregation,
and leaving the users' raw data on their device.
Approaches based on hash maps, count sketches~\cite{charikar2002finding},
or tries~\cite{gunasinghe2012sequence} require significant adaptation to be able to run on FL settings~\cite{zhu2019federated}.
Our work builds upon a very general FL framework for neural networks,
where we train a federated character-based recurrent neural network (RNN) on device.
OOV words are Monte Carlo sampled on servers during interference (details in~\ref{sec:lstmmodeling}).

While Federated Learning removes the need to upload raw user material --- here OOV words --- to the server,
the privacy risk of unintended memorization still exists (as demonstrated in~\cite{carlini2018secret}).
Such risk can be mitigated, usually with some accuracy cost, using techniques
including differential privacy~\cite{ldprlm2018}. Exploring these trade-offs is beyond the scope of this paper.

The proposed approach relies on a learned probabilistic model, and may therefore not
generate the words it is trained from faithfully. It may ``daydream" OOV words, that is, come up with
character sequences never seen in the training data. And it may not be able to regenerate some words
that would be interesting to learn.
The key to demonstrate the practicality of the approach is to answer:
(1) How frequently do daydreamed words occur?
(2) How well does the sampled distribution represent the true word frequencies in the dataset?
In response to these questions, our contributions include the following:

1. We train the proposed LSTM model on a public Reddit comments dataset with a
simulated FL environment.
The Reddit dataset contains user ID information for
each entry (user's comments or posts), which can be used to mimic the process
of FL to learn from each client's local data.
The simulated FL model is able to achieve
$90.56\%$ precision and $81.22\%$ recall for top ${10^5}$ unique words,
based on a total number of ${10^8}$ parallel independent samplings.

2. We show the feasibility of training LSTM models from daily Gboard user data
with real, on-device, FL settings.
The FL model is able to reach $55.8\%$ character-level top-3 prediction accuracy and 2.35 cross entropy on users' on-device data.
We further show that the top sampled words are
very meaningful and are able to capture words we know to be trending in the news at the time of the experiments.

\section{Method}
\subsection{LSTM Modeling}
\label{sec:lstmmodeling}
LSTM models~\cite{hochreiter1997long} have been successfully used in a variety of sequence processing tasks.
In this work we use a variant of LSTM with a Coupled Input and
Forget Gate (CIFG)~\cite{greff2017lstm}, peephole connections~\cite{gers2000recurrent} and a projection layer~\cite{sak2014long}.
CIFG couples the forget and input decisions together, thereby reducing the
number of parameters by $25\%$.
The projection layer that is located at the output gate generates the hidden state $h_t$ to reduce the dimension and speed-up training.
Peephole connections let the gate layers look at the cell state.
We use multi-layer LSTMs~\cite{sutskever2014sequence} to increase the representation power of the model.
The loss function of the LSTM model is computed as the cross entropy (CE)
between the predicted character distribution $\widehat{y}_i^{t}$ at each step $t$ and a one-hot encoding of the current
true character $y_i^t$, summed over the words in the dataset.

During the inference stage, the sampling is based on the chain rule of probability,
\begin{equation}
\text{Pr}(x_{0:T-1})=\prod_{t=0}^{T-1}\text{Pr}(x_t|x_{0:t-1}),
\label{eq:sequence}
\end{equation}
where $x_{0:T-1}$ is a sequence with arbitrary length $T$.
RNNs estimate $\text{Pr}(x_t|x_{0:t-1})$ as
\begin{equation}
P_{f_\theta}(x_t|x_{0:t-1})=P_r(x_t|f_{\theta}(x_{0:t-1})),
\label{eq:condprob}
\end{equation}
where $f_\theta$ can be perceived as a function mapping input sequence $x_{0:t-1}$ to the hidden state of the RNN.
The sampling process starts with multiple threads, each beginning with start of the word token.
At step $t$ each thread generates a random index based on $P_{f_\theta}(x_t|x_{0:t-1})$.
This is done iteratively until it hits end of the word tokens.
This parallel sampling approach avoids the dependency between each sampling thread,
which might occur in beam search or shortest path search sampling~\cite{carlini2018secret},

\begin{figure}
  \centering
    \includegraphics[width=0.5\textwidth]{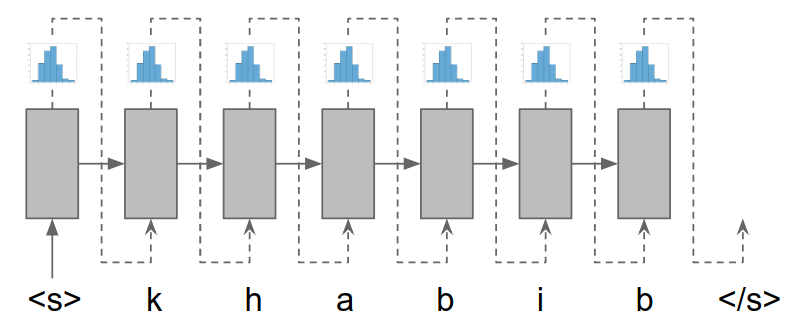}
  \caption{Monte Carlo sampling of OOV words from the LSTM model.}
  \label{fig:mcsample}
\end{figure}

\subsection{Federated Learning}
Federated learning (FL) is a new paradigm of machine learning that keeps training data localized
on mobile devices and never collects them centrally.
FL is shown to be robust to unbalanced or non-IID (independent and identically distributed) data
distributions~\cite{mcmahan2016communication, konevcny2016federated}.
It learns a shared model by aggregating locally-computed gradients.
FL is especially useful for OOV word learning, since OOV words typically include sensitive user content.
FL avoids the need for transmitting and storing such content on centralized servers.
FL can also be combined with privacy-preserving techniques like secure aggregation~\cite{BonawitzIKMMPRS16} and differential
privacy~\cite{ldprlm2018} to offer stronger privacy guarantees to users.

We use the \texttt{FederatedAveraging} algorithm
presented in~\cite{mcmahan2016communication} to combine
client updates $w_{t+1}^k\leftarrow \text{ClientUpdate}(k,w_t)$ after each round ${t+1}$ of local training to
produce a new global round with weight $w_{t+1}\leftarrow \sum_{k=1}^{K}\frac{n_k}{n}w_{t+1}^k$,
where $w^t$ is the global model weight at round $t$ and $w_{t+1}^k$ is the model weight for each participating device $k$,
and $n_k$ and $n$ are number of data points on client $k$ and the total sum of all $K$ devices.
Adaptive $L_2$-norm clipping is performed on each client's gradient,
as it is found to improve the robustness of model convergence.

\section{Experiments}
In this section, we show the details of our experiments in two different settings:
(1) simulated FL with the public Reddit dataset~\cite{al2016conversational},
(2) on-device FL where user text never leaves their device.
In all of our experiments, we use a filter to exclude ``invalid'' OOV patterns that represent things we don't want the model to learn.
The filtering excludes words that:
(1) start with non-ASCII alphabetical characters (mostly emoji),
(2) contain numbers (street or telephone numbers),
(3) contain repetitive patterns like ``hahaha" or ``yesss",
(4) are no longer than 2 in length.
In both simulated and on-device FL setting, training and evaluation are defined by separate computation tasks that sample users' data independently.

\subsection{Evaluation metrics}
In OOV word learning, we are interested in how many words are either missing or daydreamed from the model sampling.
In simulated FL, we have access to the datasets and know the ground truth OOV words and their frequencies.
Thus, the quality of the model can be evaluated based on precision and recall (PR).
For the on-device FL setting, it is not possible to compute PR since users' raw input data is inaccessible by design.
We show that the model is able to converge to good CE loss and top-K character-level prediction accuracy.
Unlike PR, CE and accuracy does not need computationally-intensive sampling and can be computed on the fly during training.

\subsection{Model parameters}
Table~\ref{table:hyperpar} shows three different model hyper-parameters used in our federated experiments.
$N_r$, $\eta$, $m$, and $B_s$ refers to number of RNN layers, server side learning rate, momentum, and batch size, respectively.
$\text{FL}^\text{SGD}_\text{S}$ and $\text{FL}^\text{SGD}_\text{L}$ applies standard SGD without momentum or clipping.
They vary in the LSTM model architectures, where model $\text{FL}^\text{SGD}_\text{S}$ contains $216$K parameters and $\text{FL}^\text{SGD}_\text{L}$ contains $758$K parameters.
$\text{FL}^\text{M}_\text{L}$ has the same model architecture as $\text{FL}^\text{SGD}_\text{L}$ and further applies adaptive gradient clipping,
combined with Nesterov accelerated gradient~\cite{nesterov1983method} and a momentum hyper-parameter of $0.9$.
Unlike server-based training, FL uses a client-side learning rate $\eta_\text{client}$ with local min-batch update,
in addition to the server-side learning rate $\eta$.
$\text{FL}^\text{M}_\text{L}$ converges with $\eta_\text{client}=0.5$, while $\text{FL}^\text{SGD}_\text{S}$ and $\text{FL}^\text{SGD}_\text{L}$ diverges with such a high value.

\begin{table}[]
\centering
\begin{tabular}{c|ccccc}
                 & $\text{FL}^\text{SGD}_\text{S}$ & $\text{FL}^\text{SGD}_\text{L}$ & $\text{FL}^\text{M}_\text{L}$     \\ \hline
$m$              & 0.0  & 0.0  & 0.9      \\
$B_s$            & 64   & 64   & 64       \\
$S$              & 0.0  & 0.0  & $6.0^*$  \\
$\eta$           & 1.0  & 1.0  & 1.0      \\
$\eta_{client}$  & 0.1  & 0.1  & 0.5      \\
$N_r$            & 2    & 3    & 3        \\
$N$              & 256  & 256  & 256      \\
$d$              & 16   & 128  & 128      \\
$N_p$            & 64   & 128  & 128
\end{tabular}
\caption{Hyper-parameters for three different FL settings.
*Here adaptive clipping with a 0.99 percentile ratio is used together with server-side L2 clipping.}
\label{table:hyperpar}
\end{table}

\subsection{Federated learning settings}
For both simulated and on-device FL, $200$ client updates are required to close each round.
For each training round we set number of local epochs as $1$.

\subsubsection{Simulated FL on Reddit data}
The Reddit conversation corpus is a publicly-accessible and fairly large dataset.
It includes diverse topics from 300,000 sub-forums~\cite{al2016conversational}.
The data are organized as ``Reddit posts'', where users can comment on each other's comments indefinitely and user IDs are kept.
The data contain 133 million posts from 326 thousand different sub-forums, consisting of 2.1 billion comments.
Unlike Twitter, Reddit message sizes are unlimited.
Similar to work in~\cite{al2016conversational}, Reddit posts that have more than 1000 comments are excluded.
The final filtered data used for FL simulation contain $492$ million comments coming from $763$ thousand unique users.
There are $259$ million filtered OOV words, among which $19$ million are unique.
As user-tagged data is needed to do FL experiments,
Reddit posts are sharded in FL simulations based on user ID to mimic the local client cache scenario.

\begin{figure}
  \centering
    \includegraphics[width=0.5\textwidth]{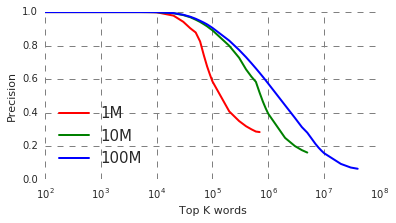}
  \caption{Precision vs. top-$K$ uniquely sampled words in simulated FL experiments.}
  \label{fig:precision_rate}
  \centering
    \includegraphics[width=0.5\textwidth]{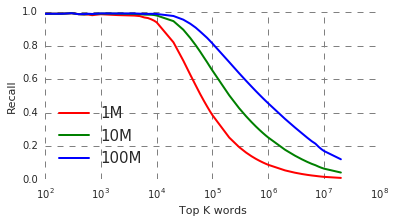}
  \caption{Recall vs. top-$K$ unique words from ground truth in simulated FL experiments.}
  \label{fig:recall_rate}
\end{figure}

\subsubsection{FL on client device data}

In the on-device FL setting, the original raw data content is not accessible for human inspection since it
remains stored in local caches on client devices.
In order to participate in a round of FL, client devices must:
(1) have at least 2G of memory,
(2) be charging,
(3) be connected to an un-metered network,
and (4) be idle.
In this study we experiment FL on three languages:
(1) American English (en\_US),
(2) Brazilian Portuguese (pt\_BR),
and (3) Indonesian (in\_ID).
A separate FL model is trained specifically on devices located in each region for each language.
Although ground truth OOV words are not accessible in the on-device setting,
the model can still be evaluated on a character-level metric like top-3 prediction accuracy or CE (i.e. given current context ``extra'' in ``extraordinary'', predict ``o'' for next character).

\section{Results}
\subsection{FL simulation on Reddit data}
During training, $\text{FL}^\text{M}_\text{L}$ converges faster and better than $\text{FL}^\text{SGD}_\text{S}$ and $\text{FL}^\text{SGD}_\text{L}$ in both CE loss and prediction accuracy,
with $66.3\%$ and $1.887$ for top-3 accuracy and CE loss, respectively.
The larger model does not lead to significant gains.
Momentum and adaptive clipping lead to faster convergence and more stable performance.

Table~\ref{table:reddittopoov} shows the top 10 OOV words with their occurring probability in the Reddit dataset (left) and the generative model (right).
The model generally learns the probability of word occurrences, where the absolute value and relative rank for top words are very close to the ground truth.

\begin{table}[]
\centering
\begin{tabular}{cc|cc}
yea          & 0.0050     & yea         & 0.0057 \\
upvote       & 0.0033     & upvote      & 0.0040 \\
downvoted    & 0.0030     & downvoted   & 0.0033 \\
alot         & 0.0026     & alot        & 0.0029 \\
downvote     & 0.0023     & downvote    & 0.0026 \\
downvotes    & 0.0018     & downvotes   & 0.0022 \\
upvotes      & 0.0016     & upvotes     & 0.0021 \\
wp-content   & 0.0016     & op's        & 0.0019 \\
op's         & 0.0015     & wp-content  & 0.0017 \\
restrict\_sr & 0.0014     & redditors   & 0.0016
\end{tabular}
\caption{Top 10 OOV words and their probabilities from ground truth (left) vs. samples (right) from simulated federated model $\text{FL}^\text{M}_\text{L}$ trained on Reddit data}
\label{table:reddittopoov}
\end{table}



In figures~\ref{fig:precision_rate} and~\ref{fig:recall_rate},
PR are computed using the model checkpoint of $\text{FL}^\text{M}_\text{L}$ with ${10^8}$ after 3000 rounds, giving $90.56\%$ and $81.22\%$, respectively.
PR rate is plotted against the top $K$ unique words (x-axis).
Curves shown in red, green, and blue represent ${10}^{6}$, ${10}^{7}$ and ${10}^{8}$
number of independent samplings, respectively.
Both the PR rate improve significantly when the amount of sampling is increased from ${10}^{6}$ to ${10}^{8}$.
We also observe increased PR over the training rounds.

\subsection{FL on local client data}
\begin{figure}
  \centering
    \includegraphics[width=0.5\textwidth]{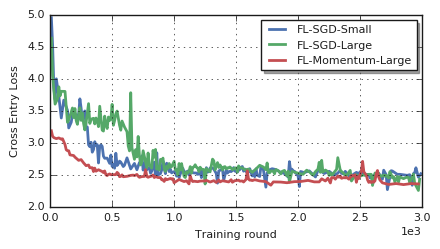}
  \caption{Cross entropy loss on live client evaluation data for three different FL settings for en\_US.}
  \label{fig:fl_ce_loss_en_us}
\end{figure}
\begin{figure}
  \centering
    \includegraphics[width=0.5\textwidth]{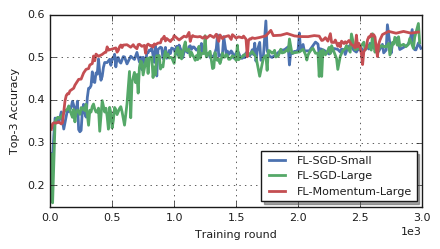}
  \caption{Top-3 Character-level prediction accuracy on live client evaluation data for three different FL settings for en\_US.}
  \label{fig:fl_accuracy_en_us}
\end{figure}
For on-device settings, all the three models converge in about $2000$ rounds over the course of about 4 days.
Figures~\ref{fig:fl_ce_loss_en_us} and~\ref{fig:fl_accuracy_en_us} compare the CE loss and top-3 prediction accuracy on evaluation data for three FL settings.
Similar to Reddit data, $\text{FL}^\text{M}_\text{L}$ converges faster and better than $\text{FL}^\text{SGD}_\text{S}$ and $\text{FL}^\text{SGD}_\text{L}$,
achieving 55.8\% and 2.35 for prediction accuracy and CE loss, respectively
(compared to $63.9\%$ and $2.01$ in training).
Experiments in pt\_BR and in\_ID shows a very similar pattern among the three settings.

Table~\ref{tab:flsamples} shows sampled OOV words in the aforementioned three languages.
Here ``\textbf{abbr.}'' is short for abbreviations.
``\textbf{slang/typo}'' refers to commonly spoken slang words and purposeful misspellings.
``\textbf{repetitive}'' refers to interjections or words that people commonly misspell in a repetitive way intentionally.
``\textbf{foreign}'' refers to words typed in a language foreign to the current language/region setting.
``\textbf{names}'' refers to trending celebrities' names.
We also observed a lot of profanity learned by the model that is not shown here.
Our future work will focus on better filtering out those OOV words, especially unintended typos.
This can be accomplished by using a manually curated desirable and undesirable OOV blacklists,
which can be updated with newer rounds of FL in an iterative process.
As we sample words, those desirable or undesirable words can be continually incorporated and made available for all users in future.
Thus the model can save more capacity to focus entirely on new words.


\begin{table}[]
\begin{tabular}{c|l|l|l}
                                                     & \textbf{en\_US} & \textbf{in\_ID} & \textbf{pt\_BR} \\ \hline
\multirow{5}{*}{\textbf{abbr.}}                      & rlly            & noh             & pqp             \\
                                                     & srry            & yws             & pfv             \\
                                                     & lmaoo           & gtw             & rlx             \\
                                                     & adip            & tlpn            & sdds            \\ \hline
\multirow{5}{*}{\textbf{slang/typo}}                      & yea             & gimana          & nois            \\
                                                     & tommorow        & duwe            & perai           \\
                                                     & gunna           & clan            & fuder           \\
                                                     & sumthin         & beb             & ein             \\ \hline
\multirow{3}{*}{\textbf{repetitive}}                 & ewwww           & siim            & tadii           \\
                                                     & hahah           & rsrs            & lahh            \\
                                                     & youu            & oww             & lohh            \\
                                                     & yeahh           & diaa            & kuyy            \\ \hline
\multirow{4}{*}{\textbf{foreign}}                    & muertos         & block           & buenas          \\
                                                     & quiero          & contract        & fake           \\
                                                     & bangaram        & cream           & the             \\ \hline
\multirow{4}{*}{\textbf{names}}                      & kavanaugh       &                 &            \\
                                                     & khabib          & N.A.            & N.A.          \\
                                                     & cardi           &                 &
\end{tabular}
\caption{Sampled OOV words from FL models in three languages (en\_US, pt\_BR, in\_ID).}
\label{tab:flsamples}
\end{table}

\section{Conclusion}
In this paper, we present a method to discover OOV words through federated learning.
The model relies on training a character-based model from which words can be generated via sampling.
Compared with traditional server-side methods, our method learns OOV words on each device and transmits the learned knowledge by aggregating gradient updates from local SGD.
We demonstrate the feasibility of this approach with simulated FL on a publicly-available corpus where we achieve $90.56\%$ precision and $81.22\%$ recall for top ${10^5}$ unique words.
We also perform live experiments with on-device data from 3 populations of Gboard users and
demonstrate that this method can learn OOV words effectively in a real-world setting.

\section*{Acknowledgments}
The authors would like to thank colleagues in the Google AI team for providing
the federated learning framework and for many helpful discussions.

\bibliography{paper}
\bibliographystyle{acl_natbib}

\appendix

\end{document}